\documentclass{article} 
\usepackage{iclr2015,times}
\usepackage{hyperref}
\usepackage{url}
\usepackage{multirow}
\usepackage{graphicx} 

\usepackage{color}

\title{Theano-based Large-Scale Visual Recognition with Multiple GPUs}
\author{Weiguang Ding \& Ruoyan Wang
\\
School of Engineering, University of Guelph\\
\texttt{\{wding, ruoyanry\}@uoguelph.ca} \\
\And
Fei Mao \\
Sharcnet, Compute Canada \\
\texttt{feimao@sharcnet.ca} 
\And
Graham Taylor 
\\
School of Engineering, University of Guelph\\
\texttt{gwtaylor@uoguelph.ca} \\
}

\iclrfinalcopy 

\begin{document}

\maketitle

\begin{abstract}
In this report, we describe a Theano-based AlexNet 
\citep{krizhevsky2012imagenet} implementation
and its naive data parallelism 
 on multiple GPUs. 
Our performance on 2 GPUs is comparable with the state-of-art 
Caffe library \citep{jia2014caffe} run on 1 GPU. To the best of our
knowledge, this is the first open-source Python-based AlexNet
implementation to-date.
\end{abstract}

\section{Introduction}

Deep neural networks have greatly impacted many application areas.  In
particular, AlexNet \citep{krizhevsky2012imagenet}, a type of
convolutional neural network \citep{lecun1998gradient} (ConvNet), has
significantly improved the performance of image classification by
winning the 2012 ImageNet Large Scale Visual Recognition Challenge
\citep{russakovsky2014imagenet} (ILSVRC 2012).  With the increasing
popularity of deep learning, many open-source frameworks have emerged
with the capability to train deep ConvNets on datasets with over 1M
examples. These include Caffe \citep{jia2014caffe}, Torch7
\citep{collobert2011torch7} and cuda-convnet
\citep{krizhevsky2012imagenet}.  However, the convenience of using
them are limited to building ``standard'' architectures.  To
experiment with brand new architectures, researchers have to derive
and implement the corresponding gradient functions in order to do
backpropagation or other types of gradient descent optimizations.

Theano \citep{bergstra2010theano,bastien2012theano}, on the other
hand, provides the automatic differentiation feature, which saves
researchers from tedious derivations and can help in avoiding errors
in such calculations. The other advantage of Theano is that it has a
huge existing user and developer base which leverages the
comprehensive scientific Python stack (102 contributors at the time of
writing).  However, there is no previously reported work of using
Theano to do large scale experiments, such as the above mentioned
ILSVRC 2012.

Here, we report a Theano-based AlexNet trained on ImageNet data%
\footnote{The code is open sourced at 
\url{https://github.com/uoguelph-mlrg/theano_alexnet}. 
In addition, a toy example is provided at 
\url{https://github.com/uoguelph-mlrg/theano_multi_gpu}.
}.
We also introduce a naive data parallelism implementation on multiple GPUs, 
to further accelerate training.

\section{Methods}
``AlexNet'' is a now a standard architecture known in the deep
learning community and often used for benchmarking. 
It contains 5 convolutional layers, 3 of which are followed by 
max pooling layers, 2 fully connected layers, and 1 softmax layer
\citep{krizhevsky2012imagenet}.
In our AlexNet implementation, we used 2 types of convolution and max pooling
operators. The 1\textsuperscript{st} is from the Pylearn2 \citep{goodfellow2013pylearn2} 
wrapper of cuda-convnet, the original implementation of AlexNet. 
The 2\textsuperscript{nd} is the recently developed Theano wrapper of 
cuDNN \citep{chetlur2014cudnn}.
We also use functions in the PyCUDA library \citep{klockner2012pycuda} 
to transfer Theano shared variables between different python processes
for two tasks: 1. loading image mini-batches into GPUs during training; and
2. exchanging weights between models trained on multiple-GPUs.

\subsection{Parallel Data Loading}

\begin{figure}[t]
\centering
\includegraphics[width=\linewidth]{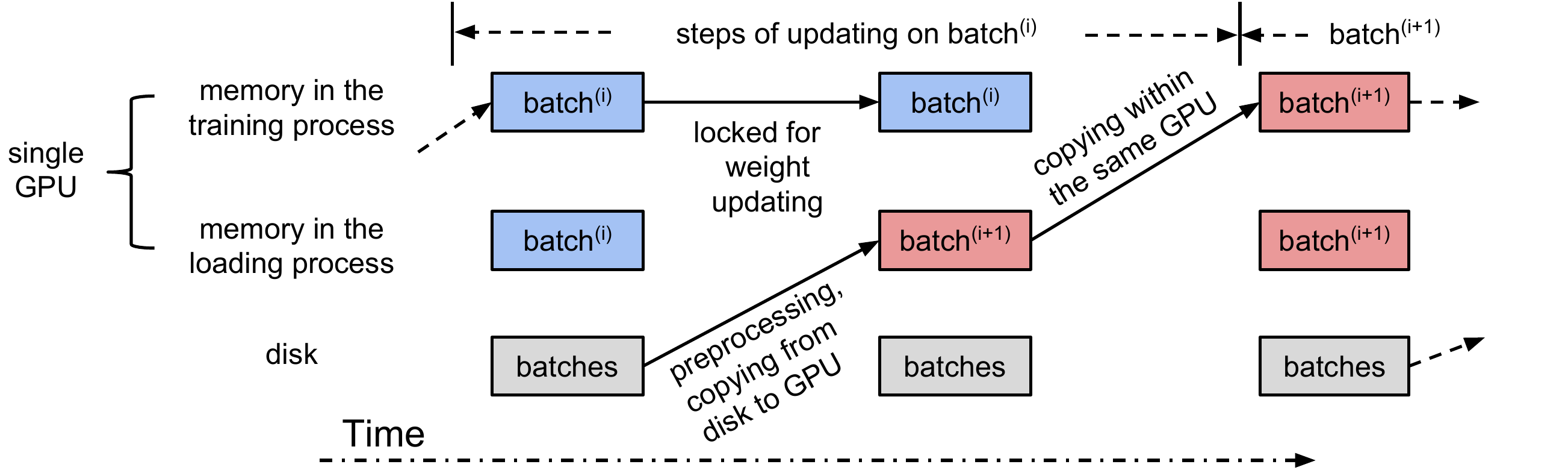}
\caption{Illustration of parallelized training and loading (1 or 2 GPUs)}
\label{fig:parallel_loading}
\end{figure}

Figure \ref{fig:parallel_loading} illustrates the process of parallelized
training and data loading. Two processes run at the same time, 
one is for training, and the other one is for loading image mini-batches.
While the training process is working on the current minibatch, 
the loading process is copying the next minibatch from disk 
to host memory, preprocessing\footnote{Preprocessing includes subtracting the mean image, 
randomly cropping and flipping images \citep{krizhevsky2012imagenet}.}
it and copying it from host memory to GPU memory.
After training on the current minibatch finishes, 
the data batch will be moved ``instantly'' from
the loading process to the training process, as they access the same GPU.

\subsection{Data parallelism}

\begin{figure}[t]
\centering
\includegraphics[width=\linewidth]{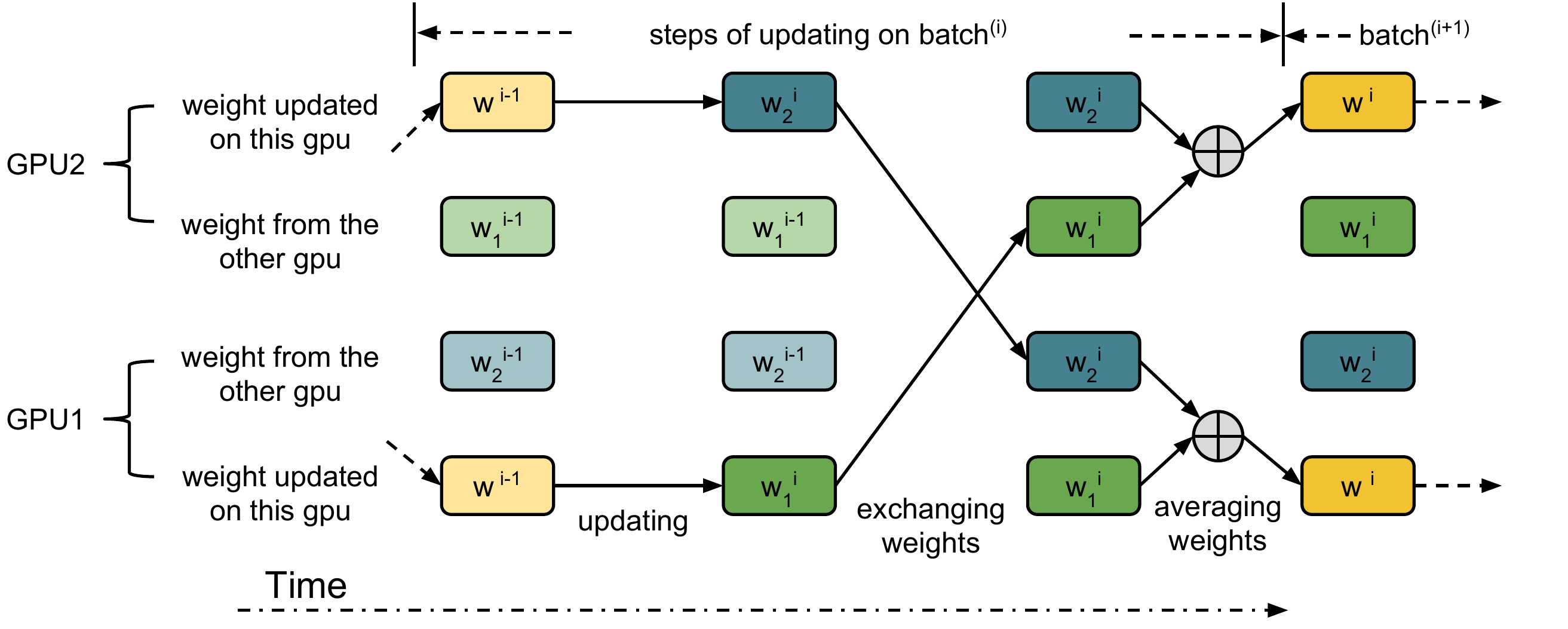}
\caption{Illustration of exchanging and averaging weights (2 GPUs)
    }
\label{fig:dual_gpu}
\end{figure}

In this implementation, 2 AlexNets are trained on 2 GPUs. 
They are initialized identically. 
At each step, they are updated on different 
minibatches respectively, and then their parameters 
(weights, biases) as well as momentum are exchanged and averaged. 

Figure \ref{fig:dual_gpu} illustrates the steps involved in training on
one minibatch. For each weight\footnote{The same operation is performed for biases and momentum.} 
matrix in the model, 
there are 2 shared variables allocated: one for updating, 
and one for storing weights copied from the other GPU. 
The shared variables for updating on 2 GPUs start the same. In the 1\textsuperscript{st} step, they are updated separately on different data batches.
In the 2\textsuperscript{nd} step, weights are exchanged between GPUs. 
In the 3\textsuperscript{rd} step, these weights (no longer the same) are averaged on both GPUs.
At this point, 2 AlexNets sharing the same parameters are ready for 
training on the next mini-batch. 

\section{Results}

\begin{table}[t]
\centering
\caption{Training time per 20 iterations (sec)}

\small
\label{tab:performance}
\begin{center}
\begin{tabular}{ccccccccc}
\hline
Parallel & \multicolumn{2}{c}{cuda-convnet} & \multicolumn{2}{c}{cuDNN-R1} & \multicolumn{2}{c}{cuDNN-R2} & \multirow{2}{*}{Caffe} &  Caffe with\\
\cline{2-7}
loading & 2-GPU & 1-GPU & 2-GPU & 1-GPU & 2-GPU & 1-GPU & & cuDNN \\
\hline
Yes & 23.39 & 39.72 &  20.58 & 34.71 & \textbf{19.72} & \textbf{32.76} &\multirow{2}{*}{26.26} & \multirow{2}{*}{20.25} \\
No & 28.92 & 49.11 & 27.31 & 45.45 & 26.23 & 43.52 & & \\
\hline
\end{tabular}
\end{center}
\end{table}

Our experimental system contains 2 Intel Xeon E5-2620 CPUs 
(6-core each and 2.10GHz),
and 3 Nvidia Titan Black GPUs. 2 of the GPUs are under the same PCI-E switch
and are used for the 2-GPU implementation. We did not
use the third GPU. For the cuDNN library, we performed experiments 
on both the version of R1 and R2.

For the experiments on a single GPU, we used batch size 256.
Equivalently, we used batch size 128 for experiments on 2 GPUs.
We recorded the time to train 20 batches (5,120 images) under 
different settings and compared them with Caffe\footnote{Performance of Caffe is according to 
\url{http://caffe.berkeleyvision.org/performance_hardware.html}, 
where timing information for CaffeNet is provided. 
As CaffeNet has similar structures, we consider this as a rough reference.}
in Table \ref{tab:performance}.

We can see that both parallel loading and data parallelism on 2 GPUs bring 
significant speed ups. 
The 2-GPU \& parallel loading implementation (cuDNN-R2) is on par with the ``Caffe with cuDNN'' implementation.

After 65 epochs of training, the top-1 class validation error rate is 42.6\%, 
and the top-5 error rate is 19.9\%, 
without the intensity and illumination data augmentation
\footnote{The pretrained parameters are available for downloading at 
\url{https://github.com/uoguelph-mlrg/theano_alexnet/tree/master/pretrained/alexnet}}.
This is within 0.5\% of the results reported in the similar Caffe implementation%
\footnote{\url{https://github.com/BVLC/caffe/tree/master/models/bvlc_reference_caffenet}}.

\section{Discussion}




\subsection{Native Theano Multi-GPU Support}
Native Theano multi-GPU support is under development\footnote{\url{
https://groups.google.com/d/msg/theano-users/vtR_L0QltpE/Kp5hK1nFLtsJ}}. 
Our present implementation is a temporary work-around before its release,
and might also provide helpful communication components on top of it.  

\subsection{Related Work}
Many multi-GPU frameworks has been proposed and implemented 
\citep{yadan2013multi,zou2014mariana,paine2013gpu,krizhevsky2014one}, 
usually adopting a mixed data and model parallelism.
This report only implements the data parallelism framework,
but it could potentially, with a non-trivial amount of effort, 
be extended to incorporate model parallelism.

\subsection{Challenges in Python-based Parallelization}
The Global Interpreter Lock
\footnote{\url{https://wiki.python.org/moin/GlobalInterpreterLock}} (GIL) 
makes parallelization difficult in CPython, 
by disabling concurrent threads within one process. 
Therefore, to parallelize, it is necessary to launch multiple processes and 
communicate between these processes. 
Straightforward inter-process communication, 
using the ``multiprocessing'' module, is very slow for 2 reasons: 
1) it serializes Numpy arrays before passing between processes;
2) communication is done through host memory. 
These problems lead us to GPUDirect peer-to-peer memory copy, 
which also has many pitfalls under the multi-process setting.
For instance, there is no host-side synchronization performed with 
device-to-device memory copy even when the sync API is called
\footnote{\url{http://docs.nvidia.com/cuda/cuda-driver-api/api-sync-behavior.html}}. 
This problem is dealt with by CUDA context syncing and 
additional message communications between processes, however,
 this and similar issues are not straightforward.

\subsection{Limitations}
To use the fast peer-to-peer GPU memory copy, 
GPUs have to be under the same PCI-E switch. Otherwise, communication
has to go through the host memory which results in longer latency.
Situations involved with more GPUs are discussed in \citet{krizhevsky2014one}.

Due to our current hardware limitation, we have only proposed and
experimented with a 2-GPU implementation.  This report and the code
will be updated once experiments on more GPUs are performed.

\section*{Acknowledgments}

We acknowledge Lev Givon for giving helpful suggestions on how to use
the PyCUDA library. We also acknowledge NVIDIA for an Academic
Hardware Grant.

\bibliography{refs}

\begin{thebibliography}{14}
\providecommand{\natexlab}[1]{#1}
\providecommand{\url}[1]{\texttt{#1}}
\expandafter\ifx\csname urlstyle\endcsname\relax
  \providecommand{\doi}[1]{doi: #1}\else
  \providecommand{\doi}{doi: \begingroup \urlstyle{rm}\Url}\fi

\bibitem[Bastien et~al.(2012)Bastien, Lamblin, Pascanu, Bergstra, Goodfellow,
  Bergeron, Bouchard, Warde-Farley, and Bengio]{bastien2012theano}
Bastien, Fr{\'e}d{\'e}ric, Lamblin, Pascal, Pascanu, Razvan, Bergstra, James,
  Goodfellow, Ian, Bergeron, Arnaud, Bouchard, Nicolas, Warde-Farley, David,
  and Bengio, Yoshua.
\newblock Theano: new features and speed improvements.
\newblock \emph{arXiv preprint arXiv:1211.5590}, 2012.

\bibitem[Bergstra et~al.(2010)Bergstra, Breuleux, Bastien, Lamblin, Pascanu,
  Desjardins, Turian, Warde-Farley, and Bengio]{bergstra2010theano}
Bergstra, James, Breuleux, Olivier, Bastien, Fr{\'e}d{\'e}ric, Lamblin, Pascal,
  Pascanu, Razvan, Desjardins, Guillaume, Turian, Joseph, Warde-Farley, David,
  and Bengio, Yoshua.
\newblock Theano: a cpu and gpu math expression compiler.
\newblock In \emph{Proceedings of the Python for scientific computing
  conference (SciPy)}, volume~4, pp.\ ~3, 2010.

\bibitem[Chetlur et~al.(2014)Chetlur, Woolley, Vandermersch, Cohen, Tran,
  Catanzaro, and Shelhamer]{chetlur2014cudnn}
Chetlur, Sharan, Woolley, Cliff, Vandermersch, Philippe, Cohen, Jonathan, Tran,
  John, Catanzaro, Bryan, and Shelhamer, Evan.
\newblock cudnn: Efficient primitives for deep learning.
\newblock \emph{arXiv preprint arXiv:1410.0759}, 2014.

\bibitem[Collobert et~al.(2011)Collobert, Kavukcuoglu, and
  Farabet]{collobert2011torch7}
Collobert, Ronan, Kavukcuoglu, Koray, and Farabet, Cl{\'e}ment.
\newblock Torch7: A matlab-like environment for machine learning.
\newblock In \emph{BigLearn, NIPS Workshop}, number EPFL-CONF-192376, 2011.

\bibitem[Goodfellow et~al.(2013)Goodfellow, Warde-Farley, Lamblin, Dumoulin,
  Mirza, Pascanu, Bergstra, Bastien, and Bengio]{goodfellow2013pylearn2}
Goodfellow, Ian~J, Warde-Farley, David, Lamblin, Pascal, Dumoulin, Vincent,
  Mirza, Mehdi, Pascanu, Razvan, Bergstra, James, Bastien, Fr{\'e}d{\'e}ric,
  and Bengio, Yoshua.
\newblock Pylearn2: a machine learning research library.
\newblock \emph{arXiv preprint arXiv:1308.4214}, 2013.

\bibitem[Jia et~al.(2014)Jia, Shelhamer, Donahue, Karayev, Long, Girshick,
  Guadarrama, and Darrell]{jia2014caffe}
Jia, Yangqing, Shelhamer, Evan, Donahue, Jeff, Karayev, Sergey, Long, Jonathan,
  Girshick, Ross, Guadarrama, Sergio, and Darrell, Trevor.
\newblock Caffe: Convolutional architecture for fast feature embedding.
\newblock In \emph{Proceedings of the ACM International Conference on
  Multimedia}, pp.\  675--678. ACM, 2014.

\bibitem[Kl{\"o}ckner et~al.(2012)Kl{\"o}ckner, Pinto, Lee, Catanzaro, Ivanov,
  and Fasih]{klockner2012pycuda}
Kl{\"o}ckner, Andreas, Pinto, Nicolas, Lee, Yunsup, Catanzaro, Bryan, Ivanov,
  Paul, and Fasih, Ahmed.
\newblock Pycuda and pyopencl: A scripting-based approach to gpu run-time code
  generation.
\newblock \emph{Parallel Computing}, 38\penalty0 (3):\penalty0 157--174, 2012.

\bibitem[Krizhevsky(2014)]{krizhevsky2014one}
Krizhevsky, Alex.
\newblock One weird trick for parallelizing convolutional neural networks.
\newblock \emph{arXiv preprint arXiv:1404.5997}, 2014.

\bibitem[Krizhevsky et~al.(2012)Krizhevsky, Sutskever, and
  Hinton]{krizhevsky2012imagenet}
Krizhevsky, Alex, Sutskever, Ilya, and Hinton, Geoffrey~E.
\newblock Imagenet classification with deep convolutional neural networks.
\newblock In \emph{Advances in neural information processing systems}, pp.\
  1097--1105, 2012.

\bibitem[LeCun et~al.(1998)LeCun, Bottou, Bengio, and
  Haffner]{lecun1998gradient}
LeCun, Yann, Bottou, L{\'e}on, Bengio, Yoshua, and Haffner, Patrick.
\newblock Gradient-based learning applied to document recognition.
\newblock \emph{Proceedings of the IEEE}, 86\penalty0 (11):\penalty0
  2278--2324, 1998.

\bibitem[Paine et~al.(2013)Paine, Jin, Yang, Lin, and Huang]{paine2013gpu}
Paine, Thomas, Jin, Hailin, Yang, Jianchao, Lin, Zhe, and Huang, Thomas.
\newblock Gpu asynchronous stochastic gradient descent to speed up neural
  network training.
\newblock \emph{arXiv preprint arXiv:1312.6186}, 2013.

\bibitem[Russakovsky et~al.(2014)Russakovsky, Deng, Su, Krause, Satheesh, Ma,
  Huang, Karpathy, Khosla, Bernstein, et~al.]{russakovsky2014imagenet}
Russakovsky, Olga, Deng, Jia, Su, Hao, Krause, Jonathan, Satheesh, Sanjeev, Ma,
  Sean, Huang, Zhiheng, Karpathy, Andrej, Khosla, Aditya, Bernstein, Michael,
  et~al.
\newblock Imagenet large scale visual recognition challenge.
\newblock \emph{arXiv preprint arXiv:1409.0575}, 2014.

\bibitem[Yadan et~al.(2013)Yadan, Adams, Taigman, and Ranzato]{yadan2013multi}
Yadan, Omry, Adams, Keith, Taigman, Yaniv, and Ranzato, Marc’Aurelio.
\newblock Multi-gpu training of convnets.
\newblock \emph{arXiv preprint arXiv:1312.5853}, 2013.

\bibitem[Zou et~al.(2014)Zou, Jin, Li, Guo, Wang, and Xiao]{zou2014mariana}
Zou, Yongqiang, Jin, Xing, Li, Yi, Guo, Zhimao, Wang, Eryu, and Xiao, Bin.
\newblock Mariana: Tencent deep learning platform and its applications.
\newblock \emph{Proceedings of the VLDB Endowment}, 7\penalty0 (13), 2014.

\end{thebibliography}
\bibliographystyle{iclr2015}

\end{document}